\documentclass[10pt,twocolumn,letterpaper]{article}

\usepackage[pagenumbers]{cvpr} %

\usepackage{graphicx}
\usepackage{amsmath}
\usepackage{amssymb}
\usepackage{booktabs}
\usepackage{bbm}
\usepackage[accsupp]{axessibility}

\usepackage[pagebackref,breaklinks,colorlinks]{hyperref}

\usepackage[capitalize]{cleveref}
\crefname{section}{Sec.}{Secs.}
\Crefname{section}{Section}{Sections}
\Crefname{table}{Table}{Tables}
\crefname{table}{Tab.}{Tabs.}

\def\cvprPaperID{1138} %
\def\confName{CVPR}
\def\confYear{2023}

\usepackage{overpic}
\usepackage{enumitem} %
\usepackage{overpic} %
\usepackage{color}
\usepackage{tabularx}

\definecolor{turquoise}{cmyk}{0.65,0,0.1,0.3}
\definecolor{purple}{rgb}{0.65,0,0.65}
\definecolor{dark_green}{rgb}{0, 0.5, 0}
\definecolor{orange}{rgb}{0.8, 0.6, 0.2}
\definecolor{red}{rgb}{0.8, 0.2, 0.2}
\definecolor{darkred}{rgb}{0.6, 0.1, 0.05}
\definecolor{blueish}{rgb}{0.3, 0.3, .6}
\definecolor{light_gray}{rgb}{0.7, 0.7, .7}
\definecolor{pink}{rgb}{1, 0, 1}
\definecolor{greyblue}{rgb}{0.25, 0.25, 1}
\definecolor{awesome}{rgb}{1.0, 0.13, 0.32}

\definecolor{figred}{rgb}{0.9, 0.1, 0.1}
\definecolor{figgreen}{rgb}{0.1, 0.7, 0.1}
\definecolor{figblue}{rgb}{0.1, 0.1, 0.9}
\definecolor{figmagenta}{rgb}{0.8, 0.1, 0.8}

\usepackage{blindtext}

\renewcommand{\paragraph}[1]{\vspace{1em}\noindent\textbf{#1}}

\usepackage{gensymb}

\usepackage{graphicx}               %

\usepackage{tabu}
\usepackage{xcolor}

\begin{document}

\title{Tracking through Containers and Occluders in the Wild}

\author{Basile Van Hoorick$^1$\hspace{.3cm} Pavel Tokmakov$^2$\hspace{.3cm} Simon Stent$^3$\hspace{.3cm} Jie Li$^2$\hspace{.3cm} Carl Vondrick$^1$\\[0.10cm]
$^1$Columbia University\hspace{.4cm} $^2$Toyota Research Institute\hspace{.4cm} $^3$Woven Planet\\[0.11cm]
\href{https://tcow.cs.columbia.edu/}{tcow.cs.columbia.edu}
}

\maketitle

\begin{abstract}
\vspace{-0.1cm}
Tracking objects with persistence in cluttered and dynamic environments remains a difficult challenge for computer vision systems. In this paper, we introduce \textbf{TCOW}, a new benchmark
and model
for visual tracking through heavy occlusion and containment. We set up a task where the goal is to, given a video sequence, segment both the projected extent of the target object, as well as the surrounding container or occluder whenever one exists. To study this task, we create a mixture of synthetic and annotated real datasets to support both supervised learning and structured evaluation of model performance under various forms of task variation, such as moving or nested containment. 
We evaluate two recent transformer-based video models and find that while they can be surprisingly capable of tracking targets under certain settings of task variation, there remains a considerable performance gap before we can claim a tracking model to have acquired a true notion of object permanence.
\vspace{-0.4cm}
\end{abstract}

\section{Introduction}
\label{sec:intro}

\begin{figure}[t]
  \centering
  \includegraphics[width=\linewidth]{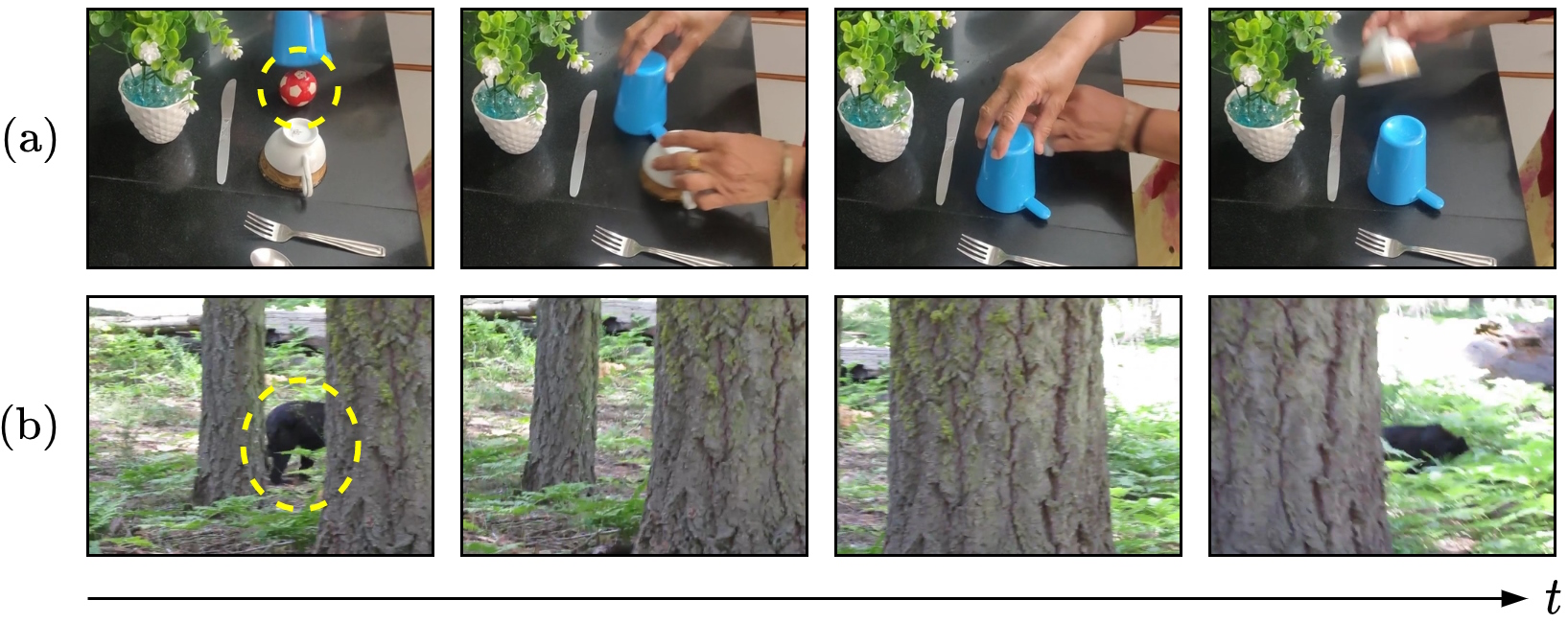}
  \caption{
  \textbf{Containment (a) and occlusion (b)} happen constantly in the real world. We introduce a novel task and dataset for evaluating the object permanence capabilities of neural networks under diverse circumstances.
  \vspace{-0.5cm}
  }
  \label{fig:teaser}
\end{figure}

The interplay between containment and occlusion can present a challenge to even the most sophisticated visual reasoning systems.
Consider the pictorial example in Figure \hyperref[fig:teaser]{1a}.
Given four frames of evidence, where is the red ball in the final frame? Could it be anywhere else? What visual evidence led you to this conclusion?

In this paper, we explore the problem of tracking and segmenting a target object as it becomes occluded or contained by other dynamic objects in a scene. This is an essential skill for a perception system to attain, as objects of interest in the real world routinely get occluded or contained.
Acquiring this skill could, for example, help a robot to better track objects around a cluttered kitchen or warehouse~\cite{bejjani2021occlusion},
or a road agent to understand traffic situations more richly~\cite{vanhoorick2022revealing}. There are also applications in augmented reality, smart cities, and assistive technology.

It has long been known that this ability, commonly referred to as object permanence, emerges early on in a child's lifetime (see \eg~\cite{baillargeon1985object, baillargeon1986representing, baillargeon1987object, spelke1990principles, baillargeon1991object, aguiar19992, spelke1994initial, aguiar2002developments, spelke2007core, piaget2013construction, spelke2013perceiving}). But how far away are computer vision systems from attaining the same?

To support the study of this question, we first propose a comprehensive benchmark video dataset of occlusion- and containment-rich scenes of multi-object interactions. These scenes are sourced from both simulation, in which ground truth masks can be perfectly synthesized, and from the real world, which we hand-annotate with object segments.
To allow for an extensive analysis of the behaviours of existing tracking systems, we ensure that our evaluation set covers a wide range of different types of containment and occlusion. For example, even though an object undergoing containment is already a highly non-trivial event, containers can move, be nested, be deformable, become occluded, and much more. Occlusion can introduce considerable uncertainty, especially when the occludee, the occluder, or the camera are in motion on top of everything else.

Using our dataset, we explore the performance of two recent state-of-the-art video transformer architectures, which we repurpose for the task of tracking and segmenting a target object through occlusion and containment in RGB video.
We show through careful quantitative and qualitative analyses that while our models achieve reasonable tracking performance in certain settings, there remains significant room for improvement in terms of reasoning about object permanence in complicated, realistic environments. 
By releasing our dataset and benchmark along with this paper, we hope to draw attention to this challenging milestone on the path toward strong spatial reasoning capabilities.

\section{Related Work}
\label{sec:rel}

\paragraph{Benchmarks for object permanence} have begun to appear in our community in recent years, but naturalistic datasets to support this study remain scarce. LA-CATER~\cite{shamsian2020learning}, based on the CATER dataset~\cite{girdhar2019cater}, is a recent example of a synthetic benchmark in which additional localization annotations for the target object were introduced when it is contained, occluded, or carried. 
More photo-realistic simulation has been applied for studying object permanence in the works of PermaTrack~\cite{tokmakov2021learning}, which uses ParallelDomain~\cite{parallel_domain}, and 4D dynamic scene completion~\cite{vanhoorick2022revealing}, which relies on CARLA~\cite{dosovitskiy2017carla}.

Notably, most prior datasets and methods focus on localizing occluded objects with a bounding box. In contrast, we focus on a more precise video object segmentation setting.
Moreover, rather than attempting to perfectly localize the invisible instance, which is not always possible in practice, we extend the setting of the problem to segmenting the \emph{occluder} instead in ambiguous scenarios. Finally, we introduce a clear distinction between containment and occlusion at the output level.

\paragraph{Object permanence methods} in computer vision were mostly studied in the context of multi-object tracking - the task of localizing all the objects from a pre-defined vocabulary with bounding boxes and associating them over time based on
identity~\cite{geiger2012we,milan2016mot16,luo2020multiple}.
As objects only need to be localized when they are visible, occlusions can be handled by simple re-association, but it has been shown that maintaining a hypothesis about the location of invisible objects can help reduce the number of identity switches~\cite{bewley2016simple}. 

To this end, most approaches rely on a simple constant velocity heuristic~\cite{yu2007multiple,breitenstein2009robust,mitzel2010multi}, which propagates the last observed location of an object with a linear motion model. It is, however, only robust when the camera is static and the object velocity does not change significantly during the occlusion (\eg because it is short). More complex, heuristic-based methods include~\cite{huang2005tracking,papadourakis2010multiple}, which localize invisible objects by modeling inter-occlusion relationships, and~\cite{grabner2010tracking} which capitalizes on the correlation between the motion of visible and invisible instances. %

More recently, several learning-based methods for localizing invisible objects have been proposed. \cite{shamsian2020learning} takes pre-computed bounded boxes for visible objects as input and passes them through a recurrent network~\cite{hochreiter1997long} that is trained to predict the bounding box for the occluded target.
In~\cite{tokmakov2021learning} and~\cite{tokmakov2021ram}, authors propose end-to-end models that are capable of localizing and associating both visible and invisible instances by capitalizing on a spatiotemporal recurrent memory~\cite{ballas2015delving, jabri2020space}.

\paragraph{Video object segmentation (VOS)} is defined as the problem of pixel-accurate separation of foreground objects from the background in videos.
In the semi-supervised VOS setting~\cite{li2013video,perazzi2016benchmark}, an algorithm is given ground truth masks for objects of interest in the first frame, and has to segment and track them for the rest of the video.

Most existing methods focus on accurately capturing object boundaries and visual appearance rather than modeling complex spatiotemporal phenomena such as object permanence.
In particular, the earliest learning-based methods~\cite{caelles2017one,khoreva2019lucid,xiao2018monet} pre-trained a CNN for binary object segmentation on static image datasets, such as COCO~\cite{lin2014microsoft}, and then separately fine-tuned this model on the first frame of the test video for each instance. Evaluating the resulting network on the remaining frames yields fairly strong accuracy, outperforming earlier, heuristic-based methods~\cite{avinash2014seamseg, fan2015jumpcut,grundmann2010efficient}. However, this approach remains computationally expensive and is not robust to appearance changes, let alone occlusion.

These limitations were later addressed in~\cite{Yang2018osmn,chen2018blazingly,hu2018videomatch}, which replace expensive fine-tuning with cheap matching, and in~\cite{voigtlaender2017online,perazzi2017learning,luiten2018premvos}, where online adaptation mechanisms are introduced for modeling the appearance of the target. More recently, memory-based models have become the mainstream approach for video object segmentation~\cite{oh2018fast,oh2019video,seong2020kernelized,yang2021collaborative,yang2021associating,cheng2022xmem}.
Generally speaking, these methods store feature maps of previous frames together with predicted instance masks in memory.
They then retrieve the closest patch with its corresponding label for every patch in the current frame to compute a segmentation. 

While these approaches demonstrate impressive performance on existing benchmarks for tracking visible objects, their reliance on visual appearance-based matching means that they cannot segment what they cannot see.
In this work, we extend the traditional VOS setting to include segmenting (the occluders and containers of) fully invisible objects, as well as amodally completing partially visible ones \cite{zhan2022tri}.
We then evaluate the state-of-the-art AOT approach~\cite{yang2021associating} and demonstrate that it indeed fails in this challenging scenario. Finally, we propose a simple modification of TimeSFormer, a transformer for video~\cite{vaswani2017, dosovitskiy2020image, bertasius2021}, to localize both visible and invisible objects, as well as distinguish containment from occlusion.

\paragraph{Sim2real.} 
Leveraging simulated data in machine learning has been essential because real-world data with exhaustive annotation is expensive to scale, or even impossible to acquire.
Promising synthetic generators and datasets have been proposed to support various tasks in different domains, including CARLA~\cite{dosovitskiy2017carla} and ParallelDomain~\cite{parallel_domain} for scene analysis and behavior understanding in autonomous driving, Flying Chairs~\cite{DFIB15} and Sintel~\cite{butler2012naturalistic} for optical flow, and ThreeDWorld~\cite{gan2020threedworld} and Kubric~\cite{greff2022kubric} for a wide variety of perception tasks in general scenes.
We observe a wide variety of data efficiency and sim2real gaps on different tasks.
For example, high generalizability can be observed in low-level feature tasks such as optical flow~\cite{teed2020raft}.
On the other hand, for tasks that involve more semantic or global context, synthetic data usually presents a more significant domain gap when transferred to the real world~\cite{toldo2020unsupervised, guizilini2021geometric,hsu2020progressive}.
Thus, selecting the right signal or task to learn from simulation is also critical. 

Our experimental results indicate that, without the need for any domain adaptation techniques, reasoning about object persistence by focusing on occluders and containers in simulated environments brings forth a surprisingly promising generalization capacity to the real world, although the overall performance is still below human abilities.

\begin{figure}[t]
  \centering
  \includegraphics[width=\linewidth]{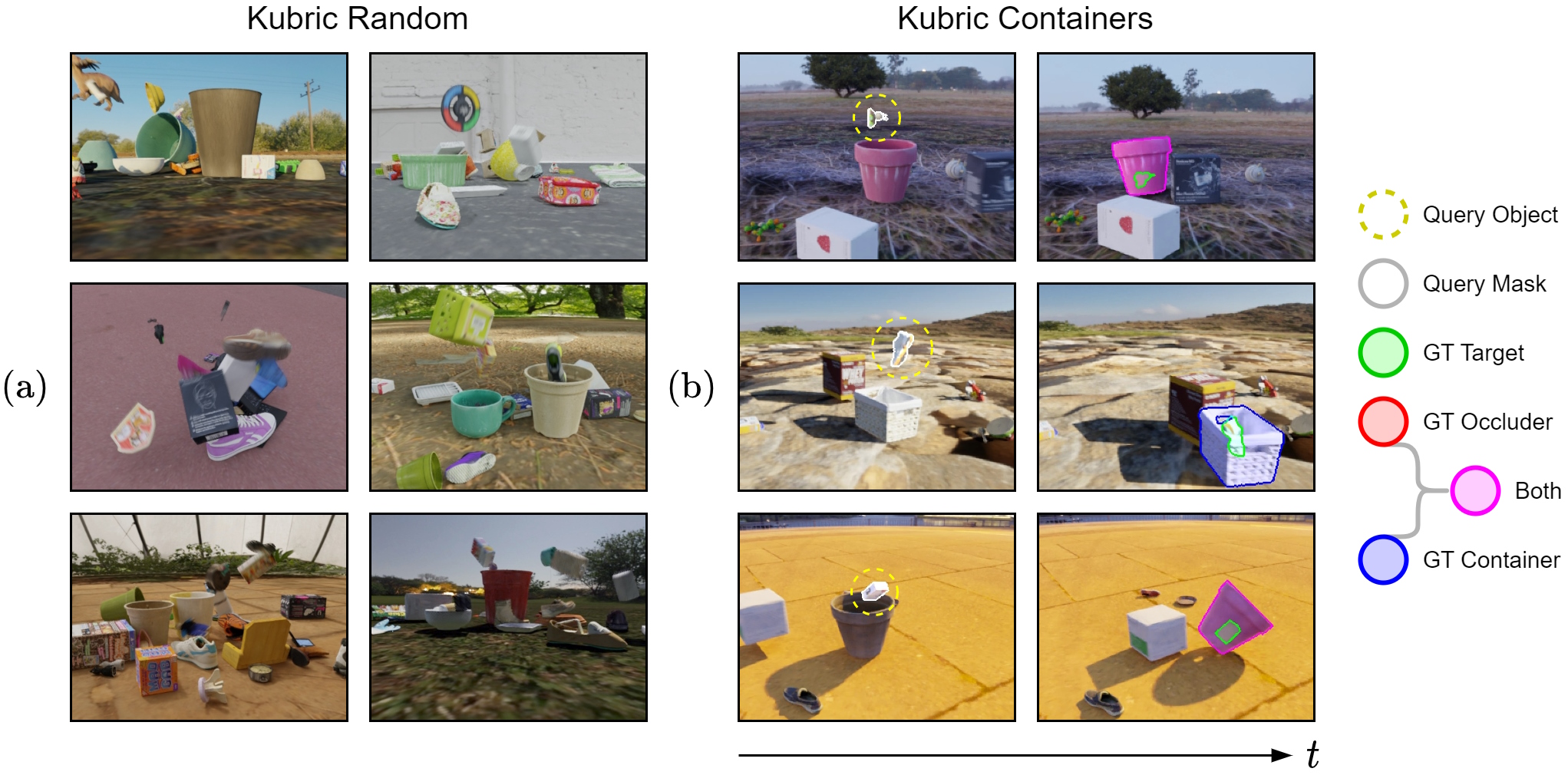}
  \caption{
  \textbf{Simulated datasets.}
  (a) We show six training examples -- these videos consist purely of randomly generated scenes in TCOW Kubric.
  (b) We show our synthetic benchmark (with annotations) where the actions are scripted -- targets fall into containers which are pushed by boxes sliding across the floor and subsequently colliding with them.
  \vspace{-0.3cm}
  }
  \label{fig:data_sim}
\end{figure}

\section{Task}
\label{sec:task}

In order to tackle object persistence thoughtfully, we propose a methodology that focuses not only on attempting to localize objects at all times, but also prompts models to explicitly consider and decide on possible containers or occluders that might be in the way.

Define $\boldsymbol x \in \mathcal{R}^{T \times H \times W \times 3}$ as the RGB-valued input video, and $\boldsymbol m_q \in \mathcal{R}^{H \times W}$ as the binary query mask, which perfectly marks the visible pixels belonging to an instance of interest in the first frame. Next, we define the function $f$, typically a neural network, whose goal is to produce segmentation masks tracking the target object and temporally propagating its mask to densely cover the rest of the video. Unlike traditional VOS settings, though somewhat similarly to \cite{zhan2022tri}, $f$ must actually predict a triplet of masks over time:
\begin{align}
\boldsymbol{\hat{y}} = f(\boldsymbol x,\boldsymbol m_q) = (\boldsymbol{\hat{m}}_t, \boldsymbol{\hat{m}}_o, \boldsymbol{\hat{m}}_c)
\label{eq:task_y}
\end{align}
Here, $\boldsymbol{\hat{m}}_t \in \mathcal{R}^{T \times H \times W}$ is the instance being tracked, $\boldsymbol{\hat{m}}_o \in \mathcal{R}^{T \times H \times W}$ is its frontmost occluder (whenever it exists), and $\boldsymbol{\hat{m}}_c \in \mathcal{R}^{T \times H \times W}$ is its outermost container (whenever it exists). Because the target object always exists \emph{somewhere} (even if located out-of-frame), the ground truth $\boldsymbol{m}_t$ is well-defined for all frames. In contrast, the occluder and container masks, $\boldsymbol{m}_o$ and $\boldsymbol{m}_c$, can be set to all-zero at moments where the target is not occluded or contained.

The triplet of ground truth segmentation masks $(\boldsymbol{{m}}_t, \boldsymbol{{m}}_o, \boldsymbol{{m}}_c)$ ought to fully characterize \emph{all} (\ie visible + invisible) pixels of their respective objects, as if X-ray goggles were provided from the camera's point of view.
For this task to become feasible for objects that have become completely hidden, it clearly requires $f$ to learn a notion of object permanence.

However, precisely pinpointing invisible objects is not always possible in practice, which compels us to find a way of dealing with irreducible uncertainty in a principled fashion. Our solution is to ask the model to reveal \emph{which} container or occluder was responsible for enveloping or hiding a target object, leading to a more expressive and interpretable representation.
Figure \hyperref[fig:data_sim]{2b} illustrates this concept.

Finally, in order to operate in a causal (online) fashion, the predicted set of masks $\boldsymbol{\hat{y}}_t$ at any time $t \in [1, T]$ may depend only on all past input frames $\boldsymbol x_{\leq t}$ up until the present.

\subsection{Evaluation Metrics}

We report the mean IoU (Intersection over Union) score, also known as region-based segmentation similarity or Jaccard index $\mathcal{J}$~\cite{perazzi2016benchmark}. In VOS, this is a conventional measure of how well a confidence-thresholded prediction overlaps with the ground truth mask~\cite{yang2021associating, cheng2022xmem}, and thus how well the model succeeds at accurately tracking the queried object of interest throughout the video.

For a sequence of target object masks $\boldsymbol{\hat{m}}_t$, the resulting IoU $\mathcal{J}_{target}$ is averaged over all frames. For the occluder and container masks $\boldsymbol{\hat{m}}_o$ and $\boldsymbol{\hat{m}}_c$, the respective IoU values $\mathcal{J}_{occl}$ and $\mathcal{J}_{cont}$ are averaged only over those frames where an occluder or container actually exists in the video.
In terms of ground truth annotations, formal definitions as to how we determine occlusion and containment events in our framework are given in Section \ref{sec:detoc}.

When evaluating multiple clips, whereas $\mathcal{J}_{target}$ is averaged uniformly across scenes, both $\mathcal{J}_{occl}$ and $\mathcal{J}_{cont}$ are weighted-averaged according to how many samples were measured per video for each type. This ensures that challenging examples with more or longer-term occlusions will be weighted more heavily than less cluttered videos where none or only a handful of frames have an active occluder.

\begin{figure*}[t]
  \vspace{-0.07cm}
  \centering
  \includegraphics[width=\linewidth]{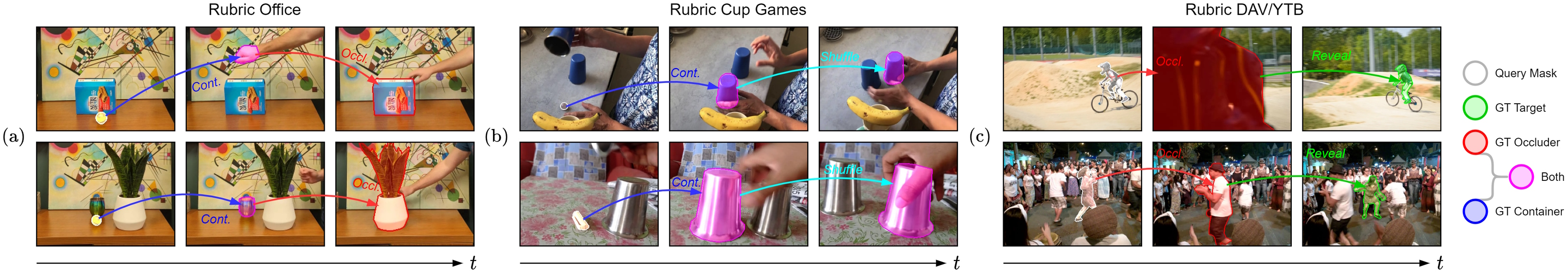}
  \caption{
  \textbf{Real-world benchmark.} We show six examples (with ground truth annotations) from Rubric Office (a), Rubric Cup Games (\ie DeepMind Perception Test~\cite{perceptiontestv1}) (b), and Rubric DAV/YTB (\ie DAVIS~\cite{pont20172017} and YouTube-VOS~\cite{xu2018youtube}) (c).
  Here, a white outline denotes the {\color{gray}query mask} in the first frame. A red outline denotes the {\color{figred}main occluder} in front of a (different) target object, and a blue outline denotes the {\color{figblue}main container} surrounding a target object, such that a magenta outline implies that one and the {\color{figmagenta}same object} is responsible for both occluding and containing the target. Finally, a green outline denotes the {\color{figgreen}target instance} itself when it re-emerges.
  }
  \label{fig:data_real}
\end{figure*}

\begin{table*}
  \centering
  \small
  \begin{tabular}{@{}lccccccc@{}}
    \toprule
    TCOW Dataset & S/R & \# Videos & \# Frames / vid. & Resolution & \# Masks / vid. & \# Cont. events / vid. & \# Occl. events / vid. \\
    \midrule
    Kubric Random & Sim & 4000 & 36 & $480 \times 360$ & 180-1188 & 0--1 & 0--4 \\
    Kubric Containers & Sim & 27 & 36 & $480 \times 360$ & 180-252 & 1 & 0--2 \\  %
    Rubric Office & Real & 32 & 150--330 & $640 \times 480$ & 3-6 & 0--2 & 0--4 \\  %
    Rubric Cup Games & Real & 14 & 308--463 & $640 \times 480$ & 4-14 & 1--3 & 0--3 \\  %
    Rubric DAV/YTB & Real & 33 & 41--180 & $640 \times 480$ & 2-9 & 0 & 0--5 \\  %
    \bottomrule
  \end{tabular}
  \caption{
  \textbf{Dataset properties.}
  TCOW consists of five parts.
  Kubric Random has a train/val/test split of 3600/200/200 scenes, and all other datasets are strictly test sets for the purpose of evaluation.
  The number of containment or occlusion events is incremented every time a potential target object enters a container or goes behind an occluder respectively.
  \vspace{-0.3cm}
  }
  \label{tab:data}
\end{table*}

\section{Datasets}
\label{sec:data}

To bring our proposed task to life,
we introduce a new collection of datasets
with the intent to facilitate both learning and evaluating object permanence. Our data is derived from synthetic sources (\textbf{TCOW Kubric}) as well as the real world (\textbf{TCOW Rubric}). %
While Kubric manifests dense, exact annotations useful for training, Rubric comprises a novel challenging benchmark for understanding object permanence in the wild. Relevant statistics are summarized in Table~\ref{tab:data}.

\subsection{Kubric}

We leverage the Kubric~\cite{greff2022kubric} simulator as the synthetic data generator for all training data, plus some evaluation videos. We modify the provided \emph{MOVi-F} template to insert containers more often, which are sourced randomly from a manually predefined list of assets within Google Scanned Objects~\cite{downs2022google}. Every scene has between 6 and 36 objects in total; roughly one-third of them are spawned in mid-air at the beginning of the video.

To construct the X-ray segmentation mask $\boldsymbol m_a \in [0,1]^{T \times H \times W \times K}$, we collect raw ground truth masks over time for all pixels of all $K$ instances separately. In addition, we study all \emph{pairs} of objects to derive any possible container-containee or occluder-occludee relationships that might emerge. Because we have access to perfect information in a simulated environment, the annotation framework described in Section \ref{sec:detoc} can be applied directly.

As shown in Figure~\ref{fig:data_sim}, we procedurally two generate versions of the TCOW Kubric dataset. First, \emph{Kubric Random} consists of a large number of cluttered scenes where the objects are spawned with independent, random velocities, thus causing various collisions and complex interactions to emerge. Occlusion and containment frequently happen by chance, encouraging neural networks to learn spatial reasoning skills and motion patterns from data.

Second, \emph{Kubric Containers} is a more constrained set of scripted videos, each of which portrays a single object falling into a container that subsequently gets pushed and displaced by a third object, \ie a moving box, that had been spawned simultaneously with a high initial horizontal velocity. Because annotations are cheap in simulation, this is the most densely labeled evaluation set.

\subsection{Rubric}

To support effective real-world evaluations, we introduce TCOW Rubric, a diverse collection of naturalistic videos depicting open-world objects experiencing containment and occlusion in various circumstances, with distinct levels of difficulty. Our data is sourced internally from videos recorded in an office space (\emph{Rubric Office}), as well as externally from DeepMind Perception Test~\cite{perceptiontestv1} (\emph{Rubric Cup Games}), DAVIS 2017~\cite{pont20172017}, and YouTube-VOS 2019~\cite{xu2018youtube} (\emph{Rubric DAV/YTB}). Figure~\ref{fig:data_real} showcases a few examples of our three real-world datasets.

\section{Labeling for Object Permanence}
\label{sec:detoc}

For evaluation and training purposes, we wish to define and distinguish occlusion and containment events when they occur.
In practice however, the state of whether an object is being occluded or being contained by another is not always clear-cut, because both concepts can be treated as a spectrum. In the following discussion, a so-called \emph{occluder-occludee} or \emph{container-containee} relationship refers to a putative object (that is not the target itself) acting as an occluder or container, \ie it is responsible for either hiding or encompassing the target instance of interest, denoted the occludee or containee respectively.
A clear formalism is required, which we first describe in the context of simulated data, where perfect information is available.

Unlike occlusion, we regard containment as being fundamentally a 3D phenomenon, because the fact that one object is inside another can generally be stated independently of camera viewpoints. In contrast, occlusions are by definition \emph{purely} a function of perspective projections to 2D images.
Hence, occlusion and containment exist as separate principles and are also calculated in different ways.

\subsection{Visible versus X-ray annotations}

Consider a dynamic scene with $K$ (not necessarily unique) objects, such that any recorded video $\boldsymbol x \in \mathcal{R}^{T \times H \times W \times 3}$ will visually depict up to $K$ objects plus the background. Define $\boldsymbol m_v \in [0, K]^{T \times H \times W}$ as an integer-valued \emph{visible segmentation mask} over time that marks the 1-based instance ID for each pixel in $\boldsymbol x$, where 0 is reserved for the background. Define $\boldsymbol m_a \in [0,1]^{T \times H \times W \times K}$ as a binary-valued \emph{X-ray segmentation mask} over time. That is, per frame $t \in [1,T]$ and per object index $k \in [1,K]$, the pixels in $\boldsymbol m_a$ are essentially boolean indicators of whether hypothetical rays emanating from the camera would hit instance $k$ at least once if it were the only object in existence.\footnote{Note that every object in isolation is treated purely as the sum of its pixels from a 2D perspective. As such, \emph{intra-}object phenomena such as self-occlusion are ignored in this paper in favor of \emph{inter-}object phenomena.} %
An arbitrary combination of objects can reside along a single ray, implying that in principle, any binary pattern is possible along the last dimension of $\boldsymbol m_a$. In particular, all values will be zero if and only if that pixel is part of the background.

\begin{figure}
  \centering
  \begin{subfigure}{\linewidth}
    \includegraphics[width=\linewidth]{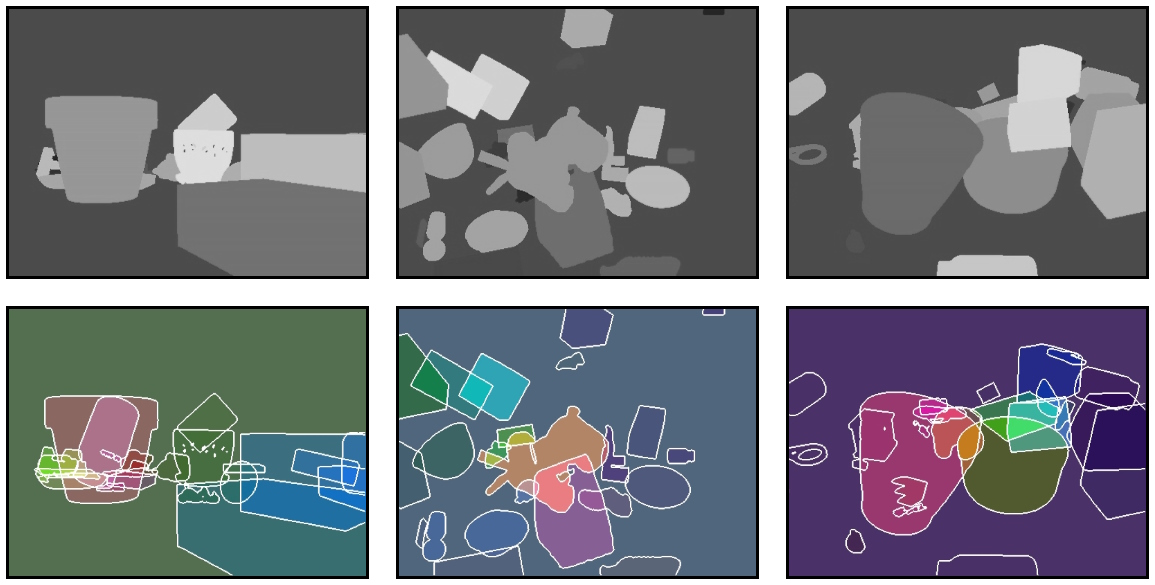}
    \caption{\textbf{Detecting occlusion} takes place by comparing visible segmentation masks $\boldsymbol m_v$ (top) with X-ray segmentation masks $\boldsymbol m_a$ (bottom -- colors are assigned randomly) to (1) identify which objects have become invisible at any point in time, and (2) for every such event, find out exactly which occluder is responsible.
    \vspace{0.1cm}
    }
    \label{fig:det_occl}
  \end{subfigure}
  \hfill
  \begin{subfigure}{\linewidth}
    \includegraphics[width=\linewidth]{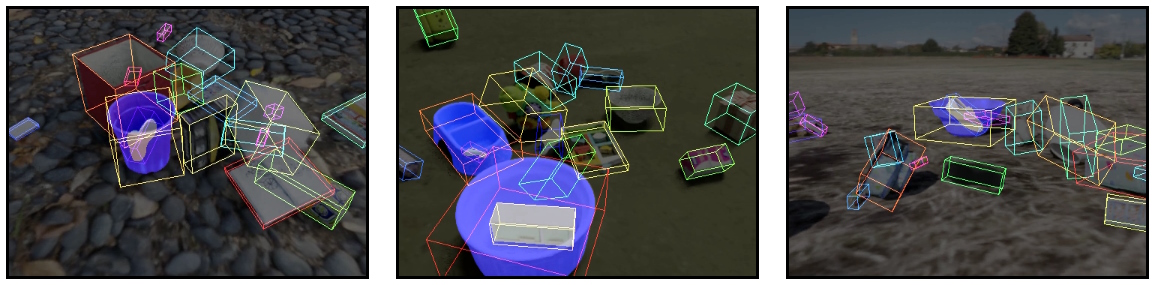}
    \caption{\textbf{Detecting containment} occurs by comparing the 3D bounding boxes between all pairs of instances, revealing when a {\color{gray}smaller object} (marked in gray) is located inside a concave {\color{blue}larger object} (marked in blue).}
    \label{fig:det_cont}
  \end{subfigure}
  \caption{
  Visualizations for understanding the methodology for gathering ground truth information with respect to inter-object interactions that pertain to object persistence.
  \vspace{-0.4cm}
  }
  \label{fig:occl_cont}
\end{figure}

\subsection{Quantifying Occlusion}

Assuming the occludee has a well-defined boundary mask, there exist varying levels of occlusion by any occluder, and we approximate this by measuring and comparing the number of \emph{visible} versus \emph{total} (\ie visible + invisible) pixels. Specifically, the \emph{occlusion fraction} (or percentage) $o_{k,t} \in [0,1]$ for instance $k$ at time $t$ is defined as follows:
\begin{align}
    o_{k,t} &= 1 - \frac{\sum_{x,y} \mathbbm{1}\left[\boldsymbol m_v(t,y,x) = k\right] } {\sum_{x,y} \boldsymbol m_a(t,y,x,k) }
\end{align}
where $\boldsymbol m_v \in [0, K]^{T \times H \times W}$ and $\boldsymbol m_a \in [0,1]^{T \times H \times W \times K}$ are the previously defined visible and X-ray segmentation masks respectively, illustrated in Figure~\ref{fig:det_occl}.

We choose a threshold of 95\%, which means that whenever the occlusion fraction satisfies $o_{k,t} \geq 0.95$, then $k$ is said to be \emph{invisible}.
Moreover, whichever object $l$ has the most (visible) pixels in front of instance $k$ is designated as its \emph{main occluder} at time $t$, and consequently populates the ground truth occluder mask $\boldsymbol m_o$ for that frame.

\subsection{Quantifying Containment}

We define $b_{k,t} \in \mathcal{R}^{3 \times 8}$ as the spatial world coordinates of the eight corners of the 3D bounding box of instance $k$ at time $t$, illustrated in Figure~\ref{fig:det_cont}.\footnote{These coordinates are embedded in a shared frame of reference with respect to the center of the Kubric scene, although the boxes themselves are not axis-aligned -- instead, they follow the canonical object frame and rotate along with its pose.} For every other object $l$, the pair-wise \emph{containment fraction} (or percentage) $c_{k,l,t}$ between a containee $k$ and its putative container $l$ is defined as follows:
\begin{align}
    c_{k,l,t} &= \frac{ |b_{k,t} \cap b_{l,t}| }{ |b_{k,t}| }
\end{align}
where $\cap$ is the geometric intersection operator, and $|b_{k,t}|$ denotes the physical 3D volume of the cuboid enveloping instance $k$.\footnote{For example, $|b_{k,t}|$ can be calculated as the absolute value of the determinant of the matrix containing the three basis vectors spanning the 3D cuboid associated with $b_{k,t}$. As for $b_{k,t} \cap b_{l,t}$ however, it is non-trivial in practice to measure volumes of the arbitrary polyhedra that may arise from intersecting two unaligned cuboids, so we instead approximate this value by densely sampling points inside $b_{k,t}$ and calculating the fraction of them that also reside within $ b_{l,t}$.}

We choose a threshold of 75\%, which means that whenever $c_{k,l,t} \geq 0.75$ (\ie more than 75\% of the target's volume is enclosed by a container $l$), then $l$ is designated as the \emph{main container} of $k$, populating the ground truth container mask $\boldsymbol m_c$ for that frame.

Rarely, we need to disambiguate multiple candidate containers $\{l_1, \ldots, l_n\}$, with $c_{k,l_i,t} \geq 0.75, \forall i \in [1,n]$. This can happen \eg in the case of nested containment if $k$ is the innermost object.
In this case, we search for whichever $l_i$ is the ``least contained'' by any other $l_j$, and is as such the \emph{outermost} container surrounding $k$ as well as all other candidates. Specifically, the main container $l_i$ of the target instance $k$ at time $t$ is defined by the solution to the optimization problem $i = \min_i \max_j c_{l_i,l_j,t}$.

\subsection{Annotations in the real world}

While it is possible (albeit expensive) to obtain visible segmentation masks $\boldsymbol m_v$ in natural videos via human annotation, accurate X-ray segmentation masks $\boldsymbol m_a$ for cluttered scenarios can typically only feasibly be retrieved via simulation due to inherent ambiguity.

For our Rubric datasets,
we first select a sparse subset of key frames depicting salient moments of interest in each video.
Then, we manually label these moments with either a target mask $\boldsymbol m_t$ when the object is fully or mostly visible, or a frontmost occluder when it is nearly completely invisible, and/or an outermost container mask when it is fully enclosed by (the convex hull of) another object.
All annotations, except where DAVIS or YouTube-VOS provided them already, were drawn by a single expert annotator by filling roughly a dozen connected line segments.

\section{Experiments}
\label{sec:exp}

In this section, we evaluate two state-of-the-art, transformer-based neural network models, in addition to several heuristics that use ground truth annotations to generate predictions. We report how well each baseline performs on both synthetic and real-world data, and analyze the main trends in success versus failure cases.

\subsection{Baseline Models}

\paragraph{AOT}:
Video object segmentation (VOS) is perhaps the most similar task to our own, so we adopt the competitive Associating Objects with Transformers (AOT) method~\cite{yang2021associating} as-is and retrain it on Kubric to teach it to track through occlusions, \ie to produce the target mask $\boldsymbol{\hat{m}}_t$ from an input video $x$ and query mask $\boldsymbol{m}_q$. We take an AOT-B checkpoint pretrained on static images (see~\cite{yang2021associating} for details), and retrain the network on the training split of Kubric Random.
However, since AOT is originally trained on YouTube-VOS and, therefore, already capable of segmenting objects in video, we also evaluate a plug-and-play variant of AOT-B without any further learning.

\paragraph{TCOW (Ours)}:
For our second baseline, we customize the competitive TimeSFormer model~\cite{bertasius2021} as backbone
to predict a triplet of masks $(\boldsymbol{\hat{m}}_t, \boldsymbol{\hat{m}}_o, \boldsymbol{\hat{m}}_c)$ given $(\boldsymbol{x}, \boldsymbol{m}_q)$, instead of a category.
We leverage its attention-based spatiotemporal context modeling capabilities and treat the output sequence as a feature map for dense video segmentation. Specifically, we ignore the classification token in favor of a linear projection from the set of embeddings after the last self-attention block back to a set of image patches of size $16 \times 16 \times 3$ that, when spatially recombined together, constitute the predictions for target, occluder, and container masks.
To ensure a fair comparison with AOT, we apply a causal mask to the attention weights inside the temporal self-attention block, to prevent information from leaking backward in time during inference.
Following~\cite{bertasius2021}, we initialize the network weights with a ViT-Base~\cite{dosovitskiy2020image} ImageNet-pretrained checkpoint, and similarly retrain it on the training split of Kubric Random.

\subsection{Baseline Heuristics}

Video instance segmentation (VIS)~\cite{yang2019video} is another closely related task. We introduce oracle baselines that have access to perfect visible instance segmentation masks and track target objects or their occluders or containers by selecting the appropriate instance from the ground truth annotations. While varying levels of thoroughness exist in imitating and repurposing expert VIS models toward object permanence, we choose the following four in order of increasing complexity:

\newcommand{\tss}{\textsuperscript}
\newcommand{\tdag}{\textdagger}

\newcommand{\tablestyle}[2]{\setlength{\tabcolsep}{#1}\renewcommand{\arraystretch}{#2}\centering\footnotesize}
\newcolumntype{Y}{>{\centering\arraybackslash}X}

\setlength{\tabcolsep}{2pt} 

\begin{table*}
\vspace{-0.07cm}
\centering
\footnotesize
\tablestyle{2.5pt}{1.05}
\begin{tabularx}{1.0\linewidth}{ll|YYYY|YYYY}
    \toprule
    Method & Training set & \multicolumn{4}{c}{Kubric Random (test set)} & \multicolumn{4}{c}{Kubric Containers} \\
    & & $\mathcal{J}_{tgt,all}$ & $\mathcal{J}_{tgt,invis}$ & $\mathcal{J}_{occl}$ & $\mathcal{J}_{cont}$ & $\mathcal{J}_{tgt,all}$ & $\mathcal{J}_{tgt,invis}$ & $\mathcal{J}_{occl}$ & $\mathcal{J}_{cont}$ \\
    \midrule
    AOT (direct plug) & Static + YouTube-VOS & 30.4 & 0.4 & 0.5\tss{*} & 1.3\tss{*} & 22.5 & 0.9 & 4.6\tss{*} & 2.3\tss{*} \\
    AOT (visible only) & Static + Kubric & 35.0 & 0.5 & 0.7\tss{*} & 1.4\tss{*} & 23.1 & 0.7 & 2.0\tss{*} & 1.9\tss{*} \\
    AOT (cartoon) & Static + Kubric (flat) & 29.8 & 5.4 & 3.7\tss{*} & 4.1\tss{*} & 20.4 & 0.9 & 4.2\tss{*} & 4.7\tss{*} \\
    AOT~\cite{yang2021associating} & Static + Kubric & 41.3 & \underline{6.8} & 5.1\tss{*} & 4.9\tss{*} & \underline{26.5} & \underline{2.5} & 6.8\tss{*} & 5.9\tss{*} \\
    \midrule
    TCOW (visible only) & ImageNet + Kubric & \underline{44.7} & 0.1 & \underline{64.6} & \underline{60.0} & 25.2 & 0.1 & \underline{73.9} & \underline{76.3} \\
    TCOW (cartoon) & ImageNet + Kubric (flat) & 31.3 & 5.6 & 30.0 & 43.6 & 21.7 & 2.3 & 26.2 & 40.1 \\
    TCOW & ImageNet + Kubric & \textbf{53.0} & \textbf{16.6} & \textbf{70.5} & \textbf{71.6} & \textbf{36.8} & \textbf{16.0} & \textbf{76.8} & \textbf{78.2} \\
    \midrule
    Copy query & - & 5.8 & 0.4 & - & - & 7.8 & 0.5 & - & - \\
    Static mask\tss{\tdag} & - & 58.3\tss{\tdag} & 10.1\tss{\tdag} & - & - & 39.3\tss{\tdag} & 10.2\tss{\tdag} & - & - \\ %
    Linear extrapolation\tss{\tdag} & - & 59.8\tss{\tdag} & 15.6\tss{\tdag} & - & - & 39.6\tss{\tdag} & 10.8\tss{\tdag} & - & - \\ %
    Jump to occluder\tss{\tdag} & - & 48.1\tss{\tdag} & - & 69.3\tss{\tdag} & - & 32.5\tss{\tdag} & - & 87.2\tss{\tdag} & - \\ %
    \bottomrule
\end{tabularx}
\caption{
\textbf{Results in TCOW Kubric (synthetic).} We report the average IOU [\%] per frame (higher is better). Our TCOW model outperforms most other baselines and ablations, and can mark both containers and occluders even more accurately than the target object itself.
\tss{*}Since AOT is incapable of outputting multiple masks for a single query instance, we compare the same prediction with all three ground truths.
\tss{\tdag}Heuristic that uses privileged information, \ie can access ground truth annotations.
}
\vspace{-0.3cm}
\label{tab:sim}
\end{table*}

\begin{figure}
  \centering
  \includegraphics[width=0.9\linewidth]{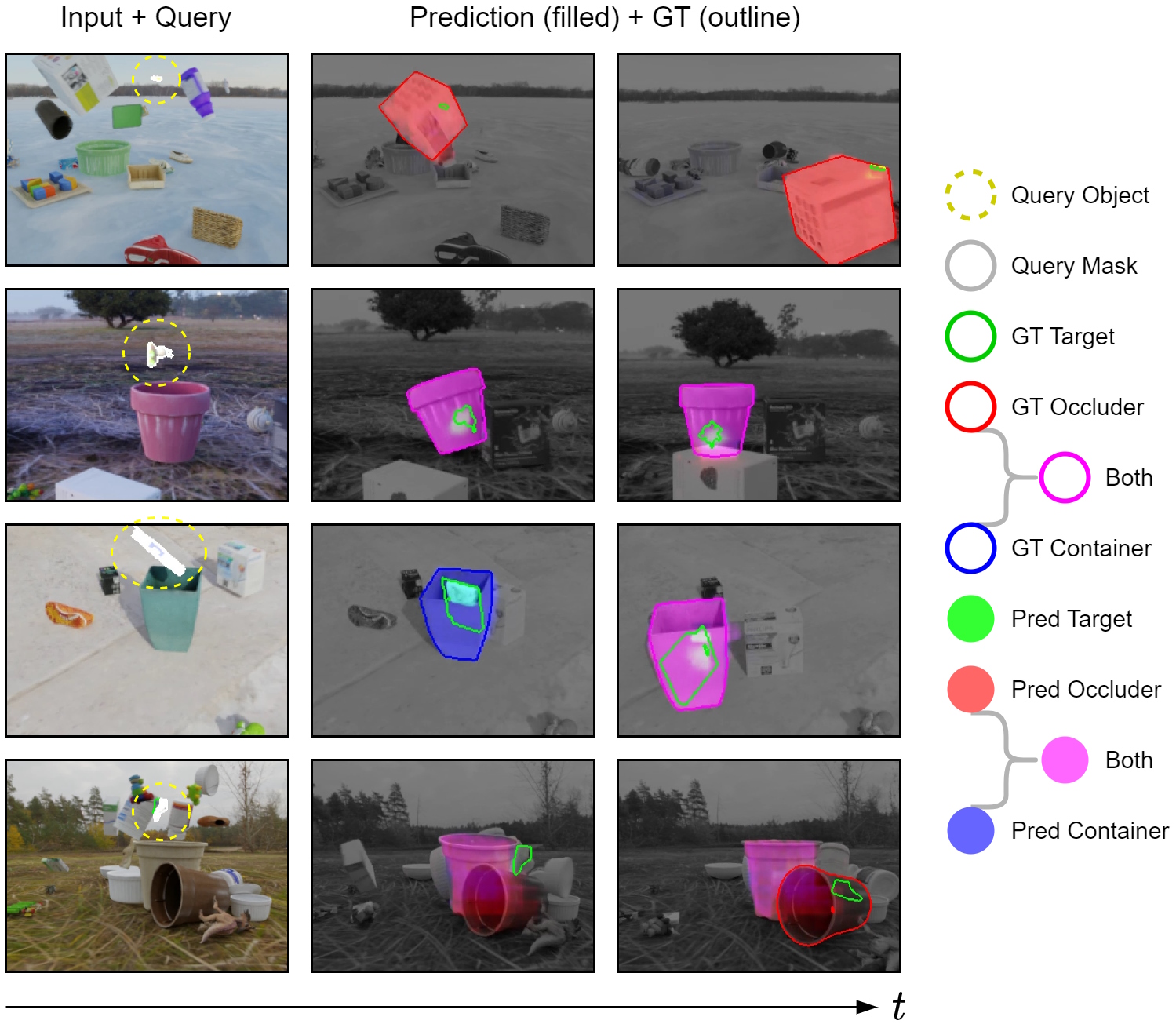}
  \caption{
  \textbf{Qualitative results for TCOW Kubric (synthetic).}
  All visualized predictions are made by the TCOW network. The first column shows the first frame along with the query mask highlighting the object we wish to track. The query object can be tiny, so we encircle it with a yellow dashed line for clarity.
  \vspace{-0.4cm}
  }
  \label{fig:res_sim}
\end{figure}

\paragraph{Copy query}: Since VOS models can see the ground truth label associated with the first frame, a simple baseline is to propagate this mask to future frames without changing it.

\paragraph{Static mask} (during occlusion): The target object is segmented perfectly whenever visible or partially occluded. During full occlusions (as defined in Section~\ref{sec:detoc}), we copy and propagate the last non-occluded ground truth X-ray mask, and hold it in that location until it re-emerges again, at which point we continue the perfect tracking routine.

\paragraph{Linear extrapolation} (during occlusion): This baseline is an extension of \emph{Static mask} that explicitly encodes and implements the constant velocity assumption that is often used as a prior in earlier works~\cite{yu2007multiple,breitenstein2009robust,mitzel2010multi,khurana2021detecting,tokmakov2021learning}. When the target instance enters a total occlusion at time $t$, its center of gravity in the two preceding frames is used to estimate an instantaneous speed vector, which is used to propagate the ground truth X-ray mask from frame $t$ until the next disocclusion occurs, at which point we return to perfect tracking.

\paragraph{Jump to occluder}: The target object is segmented perfectly to produce $\boldsymbol{\hat{m}}_t$ until the first full occlusion occurs at time $t$. Then, whichever instance has the highest number of visible pixels in front of the target's ground truth X-ray mask (\ie its main occluder) takes on the role of the object to track starting at frame $t$, filling in $\boldsymbol{\hat{m}}_o$.\footnote{Since this occluder could potentially itself also become occluded by yet another object, the described procedure may be applied recursively.} This heuristic is similar in nature to switching to tracking the nearest object when the current one has been lost, but is more powerful as it assumes knowledge of the responsible ground truth occluder.
It is the most advanced heuristic in the sense that it explicitly considers and populates the occluder mask $\boldsymbol{m}_o$, while the previous three heuristics pertain to the target mask $\boldsymbol{m}_t$ only.

\begin{table*}
\vspace{-0.07cm}
\centering
\footnotesize
\tablestyle{2.5pt}{1.05}
\begin{tabularx}{1.0\linewidth}{ll|YYY|YYY|YY}
    \toprule
    Method & Training set & \multicolumn{3}{c}{Rubric Office} & \multicolumn{3}{c}{Rubric Cup Games} & \multicolumn{2}{c}{Rubric DAV/YTB} \\
    & & $\mathcal{J}_{target}$ & $\mathcal{J}_{occl}$ & $\mathcal{J}_{cont}$ & $\mathcal{J}_{target}$ & $\mathcal{J}_{occl}$ & $\mathcal{J}_{cont}$ & $\mathcal{J}_{target}$ & $\mathcal{J}_{occl}$ \\
    \midrule
    AOT (direct plug) & Static + YouTube-VOS & \textbf{78.2} & 5.3\tss{*} & 8.2\tss{*} & 41.7 & 3.3\tss{*} & 4.3\tss{*} & \textbf{63.4} & 8.6\tss{*} \\
    AOT (visible only) & Static + Kubric & 58.0 & 4.8\tss{*} & 6.9\tss{*} & \underline{44.7} & 4.5\tss{*} & 4.6\tss{*} & 51.9 & 10.0\tss{*} \\
    AOT (cartoon) & Static + Kubric (flat) & 45.6 & 2.9\tss{*} & 3.0\tss{*} & 38.6 & 10.6\tss{*} & 9.6\tss{*} & 44.5 & 11.2\tss{*} \\
    AOT~\cite{yang2021associating} & Static + Kubric & 54.1 & 6.4\tss{*} & 8.0\tss{*} & \textbf{50.2} & 13.1\tss{*} & \underline{11.8}\tss{*} & 50.8 & 12.7\tss{*} \\
    \midrule
    TCOW (visible only) & ImageNet + Kubric & \underline{72.5} & \textbf{39.2} & \textbf{12.5} & 34.8 & \underline{27.6} & 3.5 & 51.3 & \underline{31.6} \\
    TCOW (cartoon) & ImageNet + Kubric (flat) & 35.7 & 12.1 & 7.7 & 31.9 & 8.8 & \textbf{14.3} & 22.4 & 9.2 \\
    TCOW & ImageNet + Kubric & 69.4 & \underline{30.1} & \underline{11.7} & 38.3 & \textbf{35.0} & 7.6 & \underline{52.8} & \textbf{33.4} \\
    \midrule
    Copy query & - & 12.5 & - & - & 18.6 & - & - & 15.8 & - \\
    \bottomrule
\end{tabularx}
\caption{
\textbf{Results in TCOW Rubric (real-world).} We report the average IOU [\%] per frame (higher is better). \tss{*}Since AOT is incapable of predicting multiple masks for a single query instance, we compare the same output with all three ground truths.
}
\vspace{-0.2cm}
\label{tab:real_dset}
\end{table*}

\begin{figure*}
  \centering
  \includegraphics[width=\linewidth]{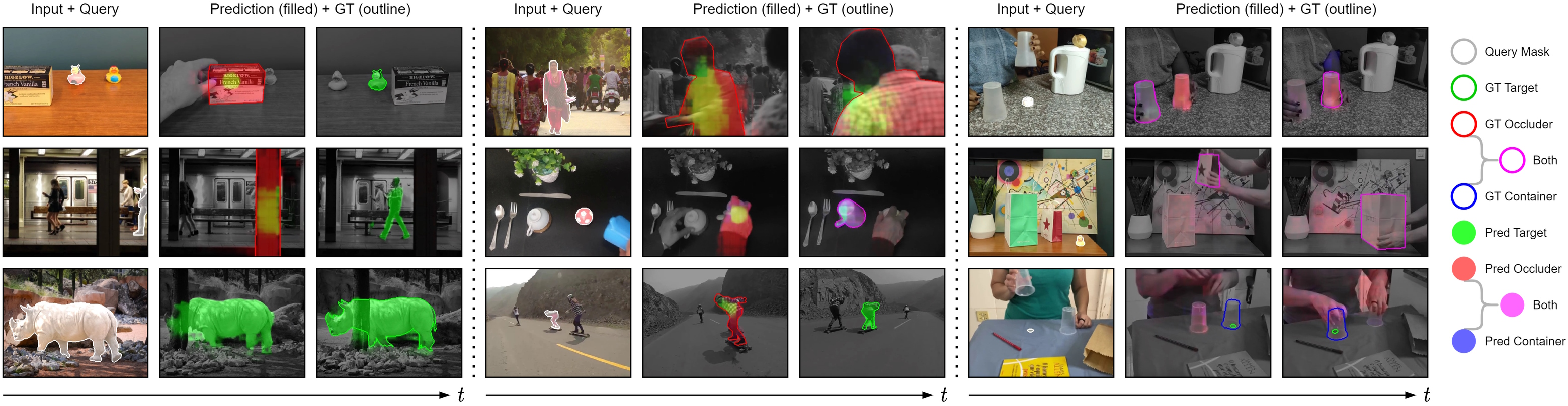}
  \caption{
  \textbf{Qualitative results for TCOW Rubric (real-world).}
  All visualized predictions are made by the TCOW network.
  We show six success cases in the left and middle columns, and three failure cases on the right.
  \vspace{-0.4cm}
  }
  \label{fig:res_real}
\end{figure*}

\subsection{Model Ablations}

Foreshadowing decent results, it is worth asking where a model's performance and generalization ability comes from in the context of object permanence. %

\paragraph{Visible pixels only}:
\emph{How important is the ability to access and directly use X-ray annotations as ground truth masks for learning to track with object permanence?} We study how the results change if we supervise models with only the visible parts of target objects and putative occluders or containers.

\paragraph{Cartoon training data}:
\emph{How important is it to ensure a faithful visual appearance of scenes when learning to track with object permanence?}
Visual realism, or the lack thereof, is often a cause for concern when working with synthetic data. While no perfect simulator exists, Kubric boasts a respectable degree of realism. Hence, we wish to examine how influential this aspect really is. To make Kubric look significantly less photorealistic, we turn off all textures by uniformly replacing all objects with unique, randomly chosen colors, as if every frame was replaced with its visible instance segmentation mask $\boldsymbol{m}_v$.

\subsection{Results}

Table \ref{tab:sim} shows quantitative results on simulated data.
$\mathcal{J}_{tgt,all}$ represents the mean Jaccard index of the target instance over all frames, but to study the localization performance of hidden objects, $\mathcal{J}_{tgt,invis}$ considers only frames where the target is fully occluded by another object.
On average, both AOT and TCOW perform somewhat similarly in terms of segmenting the target object, although TCOW shines in recognizing the correct occluder or container whenever the target becomes occluded or contained respectively. The most privileged baseline algorithm (\emph{Jump to occluder}) also works well for many cases, but is incapable of distinguishing containment from occlusion.

Figure \ref{fig:res_sim} demonstrates several examples produced by TCOW, the best-performing model, on Kubric data.
In most cases,
the main containers or occluders responsible for surrounding or concealing the target are segmented very accurately in nearly all frames.

Table \ref{tab:real_dset} shows real-world numerical results, categorized by data source.\footnote{There is no $\mathcal{J}_{tgt,invis}$ metric because fully occluded objects are never labeled in the real world; only their occluders are.}
Because AOT is the result of years of optimization by the VOS community, it boasts strong results for segmenting target objects, especially visible ones.
However, it is trained with only a relatively short context of 5 frames, which works well for conventional VOS, but seems to break down in terms of longer-term spatiotemporal reasoning, which is required for object permanence.

The decent performance of the `TCOW (visible only)' ablation suggests that it is often more fruitful to track the surrounding occluder or container of a fully hidden target object $k$ rather than to try to precisely localize $k$ at all times, which supports our task definition in Section \ref{sec:task}. Moreover, the fair performance of the ‘TCOW (cartoon)’ ablation suggests that learning the correct motion signals and occlusion/containment dynamics is important for capturing object permanence, and the remaining gap is filled by adding more realism.

Figure \ref{fig:res_real} shows representative success cases and failure cases made by the non-ablated TCOW model on real-world data. In general, this network performs surprisingly well -- for example, total occlusions involving occludees and/or occluders far outside of the training distribution are often still handled fairly correctly.
For partially occluded instances, such as the rhino on the lower left, a solid amodal completion capability is demonstrated as well.

However, there exist many Rubric videos where both models break down almost completely.
Comparing Table~\ref{tab:sim} with Table~\ref{tab:real_dset}, the quality of the occluder and container masks drop substantially when moving from synthetic to real data. Containment in particular appears to be the more difficult concept to learn robustly~\cite{hespos2001infants}.
We qualitatively observe that recursive containment, as exemplified with paper bags going inside one another in Figure~\ref{fig:res_real} (center right), is among the toughest to tackle. In fact, there is not a single such example in Rubric that is addressed satisfactorily.
Tracking objects through containment by upside-down cups that are repeatedly shuffled around also turns out to be highly demanding, especially when the cups are identical.
Lastly, videos with transparent containers present yet another failure scenario, presumably because non-opaque objects do not exist in the Kubric training data.

\section{Discussion}
\label{sec:disc}

In this work, we propose the challenging TCOW benchmark, which in its totality covers many different types of containment and occlusion, including compositions thereof. The TCOW model, based on TimeSFormer, shows promising yet lacking performance, and we believe future tracking models ought to address and resolve these scenarios more effectively.
While we have made significant strides in solving elementary base cases of occlusion and containment, object permanence as a whole remains far from being solved. We, therefore, invite and encourage the community to work on this problem.

{\small
\textbf{Acknowledgements:}
We thank Revant Teotia, Ruoshi Liu, Scott Geng, and Sruthi Sudhakar for helping record TCOW Rubric videos.
This research is based on work partially supported by the Toyota Research Institute, the NSF CAREER Award \#2046910, and the NSF Center for Smart Streetscapes (CS3) under NSF Cooperative Agreement No.\ EEC-2133516. The views and conclusions contained herein are those of the authors and should not be interpreted as necessarily representing the official policies, either expressed or implied, of the sponsors.
}

{\small
\bibliographystyle{ieee_fullname}
\bibliography{0_bib}
}

\twocolumn[
\centering
\Large
\textbf{Tracking through Containers and Occluders in the Wild} \\
\vspace{0.5em}Supplementary Material \\
\vspace{1.0em}
] %
\appendix

\section{Dataset Details}

For our training set TCOW Kubric Random, all scenes are generated based on the \emph{MOVi-F}\footnote{This is the same as \emph{MOVi-E}, but with a small degree of motion blur added to the video recorded by the virtual camera.} template code~\cite{greff2022kubric}, but with several modifications.
Backgrounds are chosen randomly from the Polyhaven HDRI collection~\cite{polyhaven}, and all objects originate from Google Scanned Objects (GSO)~\cite{downs2022google}.
Every scene spawns $s$ static objects lying on the ground, and $d$ dynamic objects falling down when the video starts. $s$ is uniformly randomly chosen between 4 and 24 (inclusive), while $d$ is uniformly randomly chosen between 2 and 12 (inclusive).

\begin{figure}
  \centering
  \includegraphics[width=\linewidth]{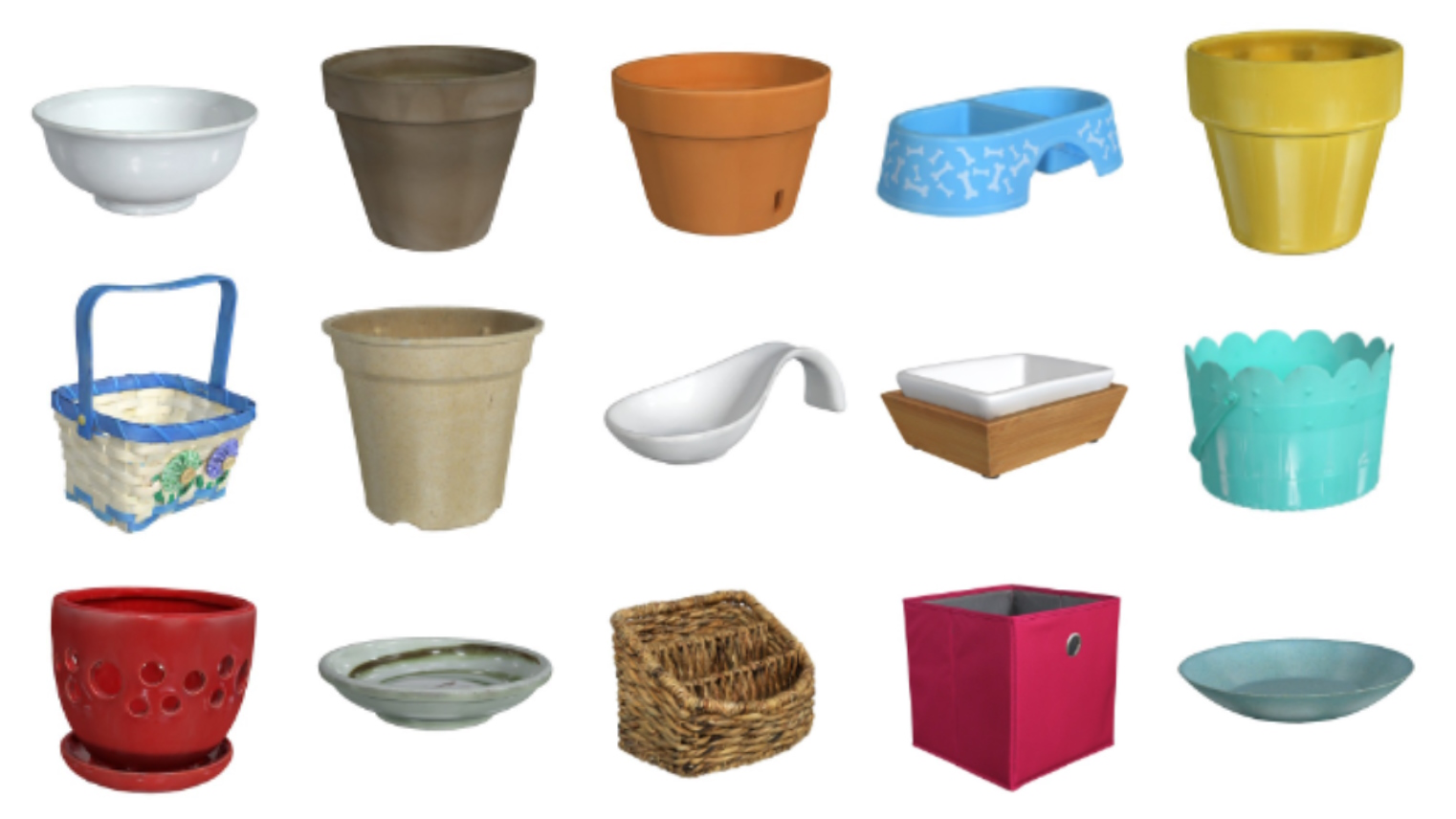}
  \caption{
  \textbf{Containers in GSO.}
  We mark roughly 11\% of the assets in Google Scanned Objects~\cite{downs2022google} to be containers, which are spawned more often than average compared to other object types in Kubric Random.
  }
  \label{fig:more_cont}
\end{figure}

In order to increase the frequency of containment, we manually scan the GSO library to designate 114 out of 1,032 GSO assets as \emph{containers}, which can be either deep or shallow.
For each scene, at least three out of the $s$ static objects must be containers, and while object sizes are chosen randomly, we also make containers slightly bigger on average. An assortment of examples is shown in Figure~\ref{fig:more_cont}.

The most time-consuming part of the simulation is generating the X-ray segmentation mask $\boldsymbol m_a \in [0,1]^{T \times H \times W \times K}$, which is used for supervision as it exposes all pixels of all $K$ instances separately over time, regardless of occlusion.
This is done by running the PyBullet physics simulation~\cite{coumans2016pybullet} once, thus letting the object interactions develop over time within the dynamic scene, then rendering the input video via Blender~\cite{blender} with all instances present, following ~\cite{greff2022kubric}. Next, we isolate each object by turning off the visibility of all other objects (they are essentially temporarily removed from existence), and rendering those videos again separately to iteratively produce one channel of $\boldsymbol m_a$ at a time.

Finally, even though the frame rate of video clips in the Rubric benchmark is variable (\ie between 4 and 30), rendering of all Kubric simulations happens at a single fixed value of 12 FPS.

To construct the Kubric Random dataset, consisting of 4,000 videos of 36 frames each with spatial dimension $480 \times 360$ along with RGB information, depth maps, and segmentation maps ($\boldsymbol m_v$ and $\boldsymbol m_a$), 256 AMD EPYC 7763 CPU cores worked for 30 days.

\begin{figure*}
  \centering
  \includegraphics[width=0.75\linewidth]{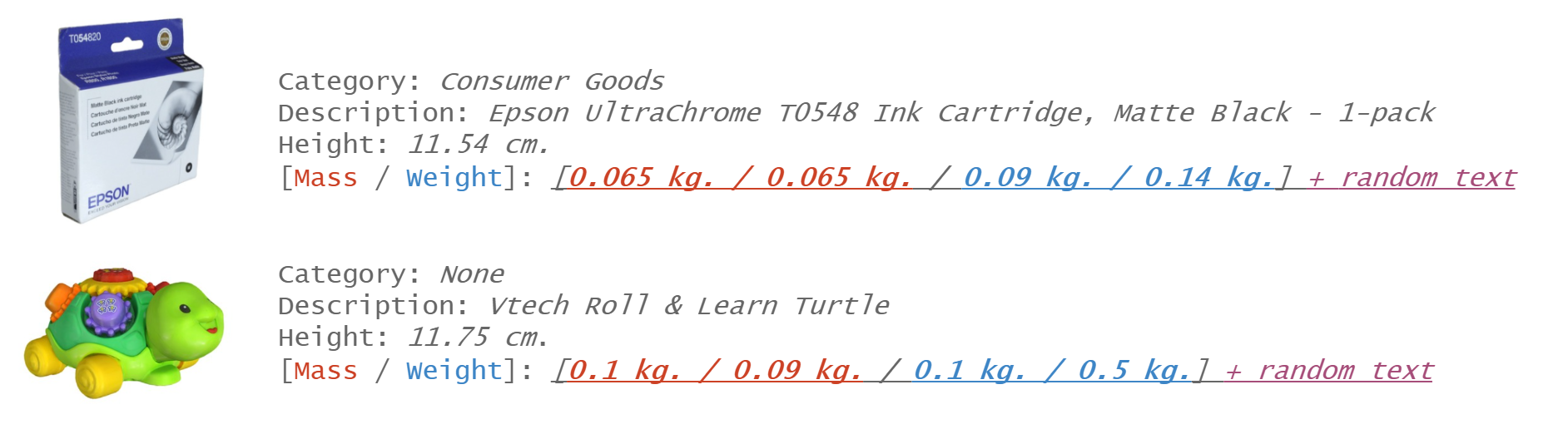}
  \caption{
  \textbf{Estimating mass for objects used in Kubric simulations.}
  We perform text completion with a large language model. Specifically, we query OpenAI GPT-3 (text-davinci-002)~\cite{brown2020language} twice for mass, twice for weight, and average the four numerical outputs after appropriate unit conversions.
  The image is shown for visualization only, and is not fed to the language model.
  The underlined text represents the four actual completion outputs made by GPT-3.
  The italic parts of the input are derived from the available metadata of each asset, and this procedure is repeated for all 1,032 GSO objects.
  }
  \label{fig:more_gpt}
\end{figure*}

\subsection{Mass Estimation}

Mass plays an important role in determining the outcome of object dynamics and interactions. While GSO provides a diverse collection of high-quality scanned 3D models for household items, physical properties such as mass and friction were not captured for many objects~\cite{downs2022google}. In Kubric \emph{MOVi-F}, a constant density assumption is therefore made by default to estimate mass from volume~\cite{greff2022kubric}. In an attempt to increase the realism of our training data, we leverage GPT-3~\cite{brown2020language} to produce rough estimates of the mass of every object in the GSO library based on its description and metadata. This is illustrated in Figure~\ref{fig:more_gpt}.
In practice, we calculate and apply the geometric mean of the original and LLM-estimated mass, because the numbers provided by GPT-3 are, qualitative speaking, not always very accurate.

\begin{figure*}
  \centering
  \includegraphics[width=\linewidth]{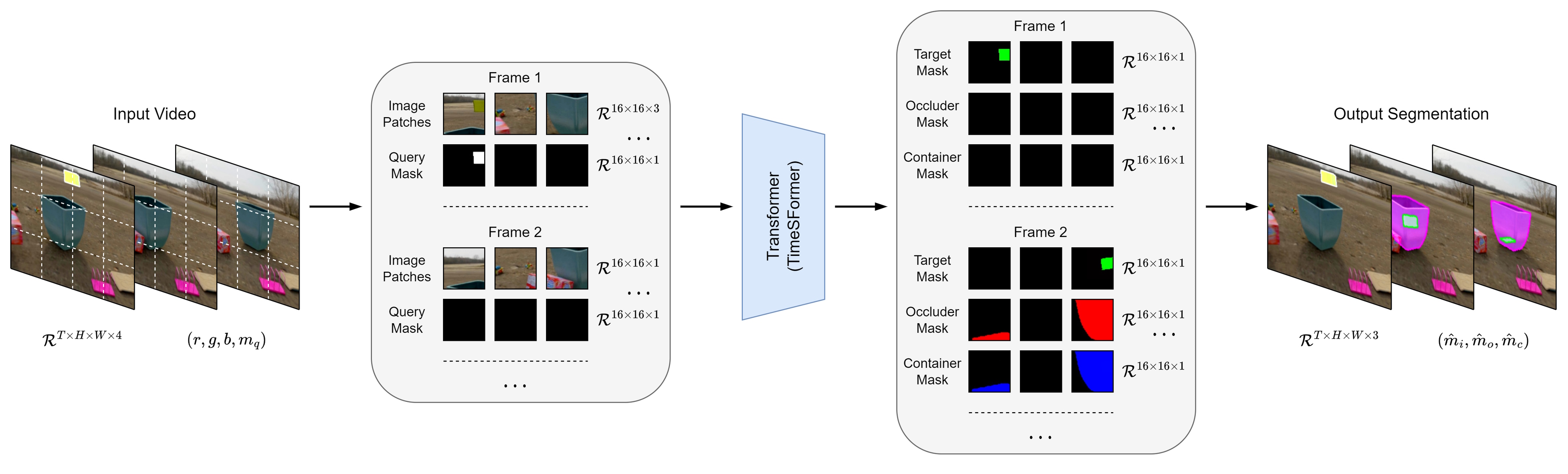}
  \caption{\textbf{TCOW architecture.} We apply the standard TimeSFormer backbone onto the input video $(\boldsymbol x, \boldsymbol m_q)$ following a spacetime divided attention scheme~\cite{bertasius2021}, but interpret the tokens after the transformer as patches for the predicted output masks. (Multiple channels belonging to the same patch are shown in separate tiles for clarity.)}
  \label{fig:arch}
\end{figure*}

\section{Network Implementation Details}

\subsection{AOT}

Since AOT is designed for VOS, we keep the entire pipeline of the AOT model intact for fairness. Following~\cite{yang2021associating}, at training time, a context window of 5 frames is fed into the model for a single training step, while at test time, the target object mask is propagated throughout the entire video clip from start to end.

\subsection{TCOW}

TCOW is a modification of the TimeSFormer network, which operates by processing a number of chunks of spacetime patches into a transformer~\cite{vaswani2017, bertasius2021}. Specifically, we concatenate the input video and the query mask along the channel axis to form $(\boldsymbol x, \boldsymbol m_q) \in \mathcal{R}^{T \times H \times W \times 4}$ (here, $\boldsymbol m_{q,t}=0$ for all $t \geq 1$ as only the first frame is labeled).
Similarly to Vision Transformer~\cite{dosovitskiy2020image}, the resulting set of frames is decomposed into $N=T \times h \times w$ small image patches of size $16 \times 16 \times 4$ each, with $h=\frac{H}{16}$, $w=\frac{W}{16}$. After a per-patch linear projection, an input sequence of $N$ embeddings
of dimensionality 768 is fed into a transformer, where we subsequently apply repeated multi-head self-attention blocks on these tokens.

The output sequence is treated as a spatiotemporal feature map for the purpose of dense video segmentation. Each element after the last attention layer
is linearly projected back to image space, resulting in a set of patches of $16 \times 16 \times 3$, where the last dimension represents the predicted triplet of masks $(\boldsymbol{\hat{m}}_t, \boldsymbol{\hat{m}}_o, \boldsymbol{\hat{m}}_c)$.
The vectors are composed in the same order as they were decomposed at the input side. The classification token is ignored and there is no pooling. A diagram is shown in Figure~\ref{fig:arch}.

\begin{figure*}
  \centering
  \includegraphics[width=\linewidth]{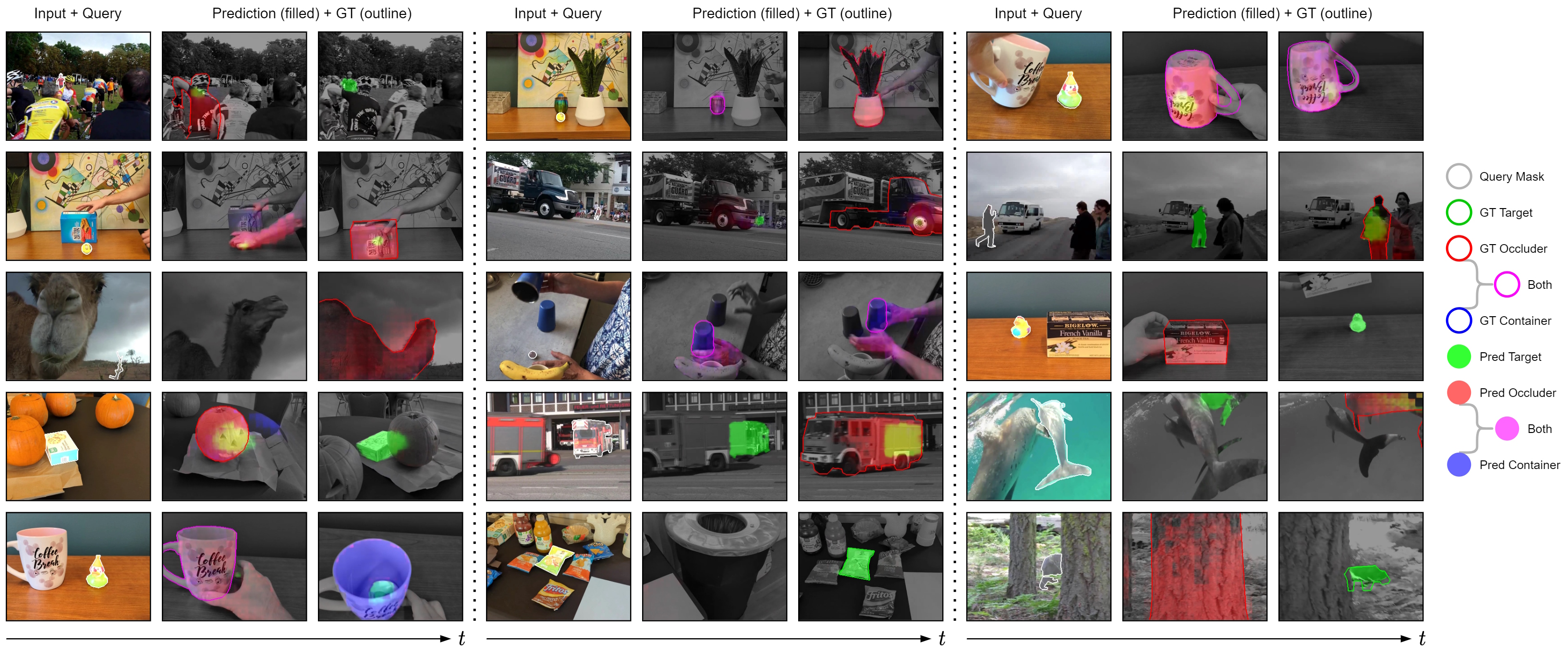}
  \caption{
  \textbf{Success cases for TCOW on Rubric.}
  All visualized predictions are made by the non-ablated TCOW network. This model performs particularly well on relatively simple cases of (total) occlusion and/or containment in the real world, despite being trained on synthetic data only. Some video clips with containers moving to a limited degree are also handled correctly (see middle center, or top right). However, more advanced examples of object permanence often result in failures, shown in Figure~\ref{fig:more_fail}, demonstrating that a lot of room for improvement remains.
  }
  \label{fig:more_succ}
\end{figure*}

\begin{figure*}
  \centering
  \includegraphics[width=\linewidth]{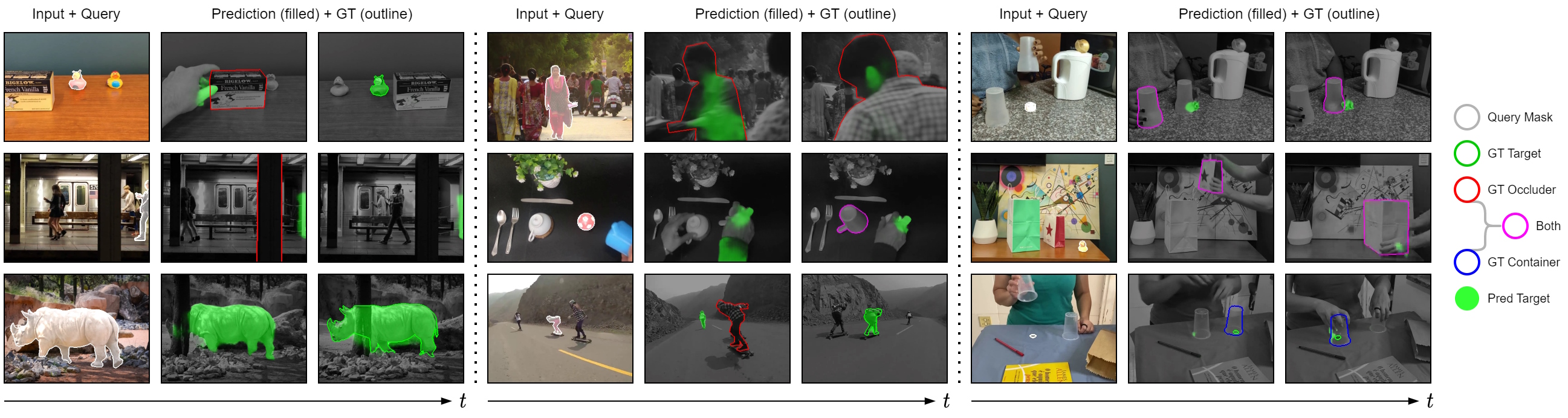}
  \caption{
  \textbf{Qualitative results for AOT on Rubric.}
  All visualized predictions are made by the non-ablated AOT network, and mirror Figure~\ref{fig:res_real} in the main text. Although all models are trained on Kubric data with X-ray supervision, AOT often loses track as soon as total occlusion happens, and tends to jump to different instances or moving parts of the video (such as hands).
  }
  \label{fig:more_base}
\end{figure*}

\begin{figure}[t]
  \centering
  \includegraphics[width=\linewidth]{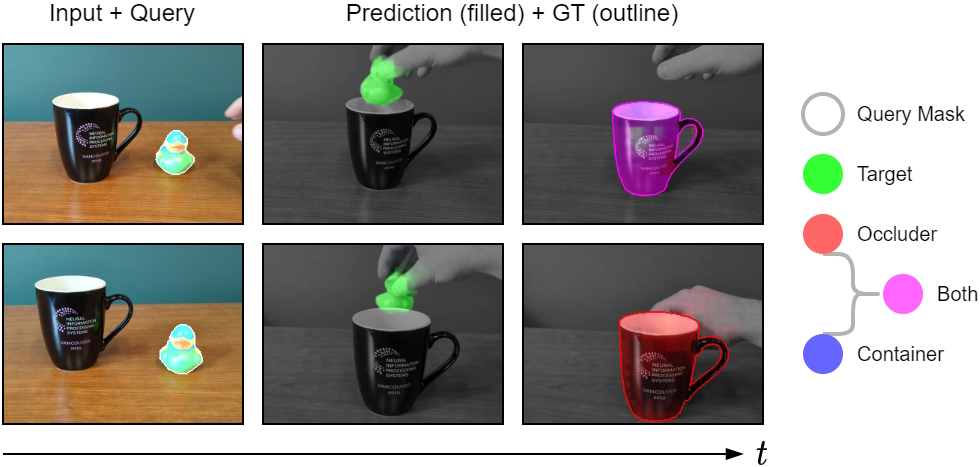}
  \caption{
  \textbf{Measuring TCOW's ability to differentiate occlusion from containment.}
  The first video is a control example where the duck is inserted \emph{inside} the mug, such that it becomes simultaneously a {\color{figmagenta}container and occluder (magenta)}. In the second video, we pretend to do the same, but actually place it \emph{behind} the mug, such that it becomes an {\color{figred}occluder (red)} only. Our TCOW model handles both cases correctly, suggesting that the learned representation is capable of spatial reasoning in a way that goes beyond just memorizing object class information (\eg a container must contain an object whenever it hides one).
  \vspace{-0.4cm}
  }
  \label{fig:confuse}
\end{figure}

\begin{figure*}
  \centering
  \includegraphics[width=\linewidth]{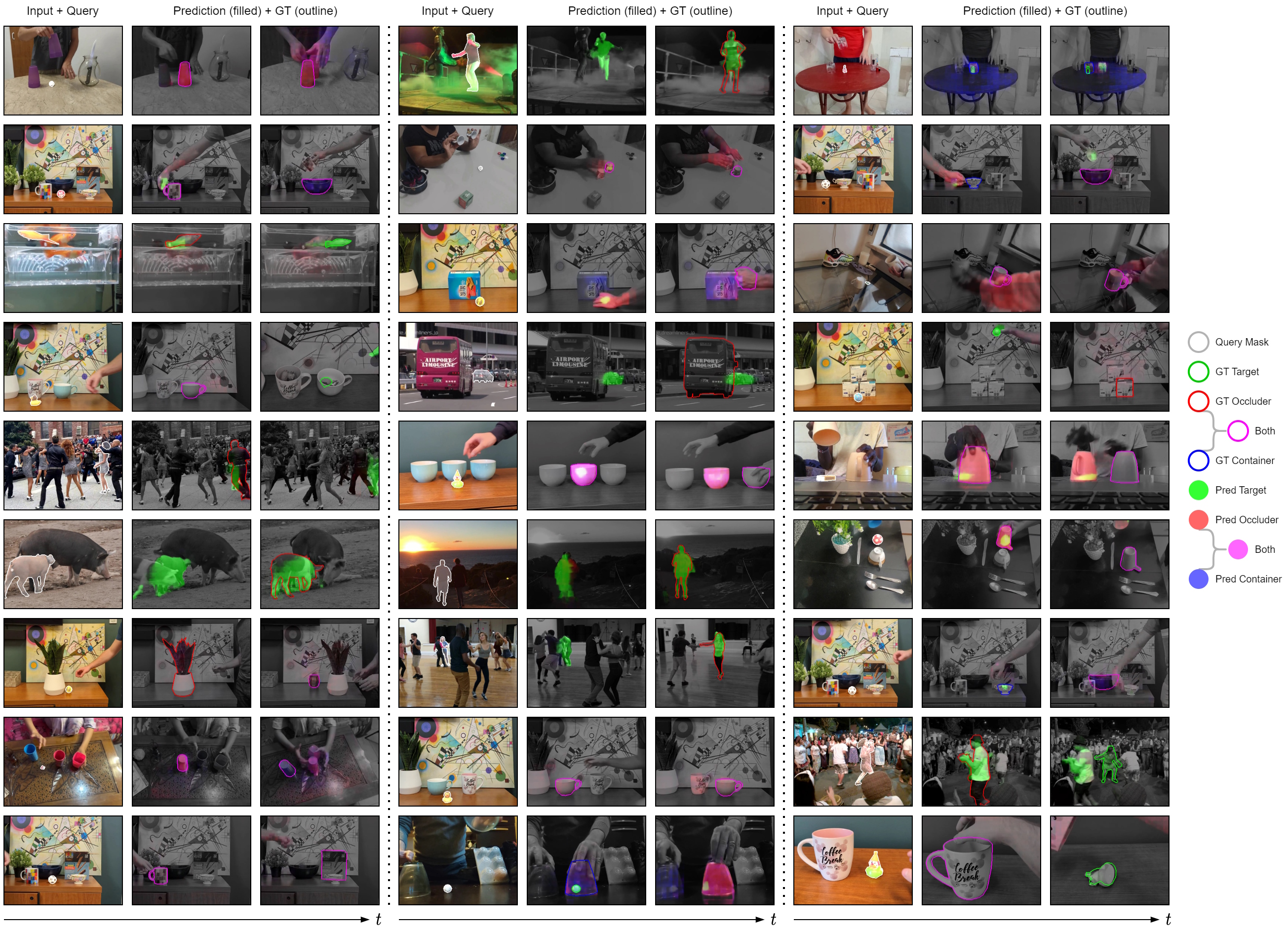}
  \caption{
  \textbf{Failure cases for TCOW on Rubric.}
  All visualized predictions are made by the non-ablated TCOW network. Multiple trends could be discerned among real-world scenarios where the model fails, which can roughly be summarized as: (1) identical containers, one of which is holding the target object, being shuffled around; (2) nested containment, \eg when a mug is placed inside a larger box; (3) the occluder and occludee are visually very similar, \eg people occluding people or animals occluding animals. By releasing this challenging benchmark to the community, we hope future work will be able to address these cases more successfully.
  }
  \label{fig:more_fail}
\end{figure*}

\subsection{Learning and Supervision}

We train the TCOW model for tracking objects through occlusion and containment by producing segmentation masks for each type.
The network $f$ (as defined in Equation~\ref{eq:task_y}) accepts a single query instance at a time, which makes a binary cross-entropy objective $\mathcal{L}_{BCE}$ between every output channel $\boldsymbol{\hat{m}}$ and its corresponding ground truth $\boldsymbol{m}$ a logical starting point. 

Since the number of frames where the target is occluded is typically smaller than the number of frames where the target is visible in our training set, we scale $\mathcal{L}_{BCE}$ by a factor $1 + (\beta - 1)o$, where $o \in [0,1]$ is the occlusion fraction.

However, inspired by~\cite{yang2021associating}, we also combine $\mathcal{L}_{BCE}$ with two additional loss terms: (1) a bootstrapped variant $\mathcal{L}_{{BCE},k}$ that focuses on a certain top fraction $k$ of pixels in each example that incur the highest individual contributions to the loss $\mathcal{L}_{BCE}$, and (2) a soft Jaccard loss $\mathcal{L}_{\mathcal{J}}$~\cite{bertels2019optimizing}.
The terms are linearly combined and weighted as follows:
\begin{align}
    \mathcal{L}_{\boldsymbol m} = (\lambda_1 \mathcal{L}_{{BCE}} +
    \lambda_2 \mathcal{L}_{{BCE},k} +
    \lambda_3 \mathcal{L}_{\mathcal{J}})(\boldsymbol{\hat{m}}, \boldsymbol{m})
\end{align}
Finally, the total objective is a weighted sum over the three different output types predicted by $f$:
\begin{align}
    \mathcal{L} = \lambda_t \mathcal{L}_{\boldsymbol m_t} +
    \lambda_o \mathcal{L}_{\boldsymbol m_o} +
    \lambda_c \mathcal{L}_{\boldsymbol m_c}
\end{align}
where $\mathcal{L}_{\boldsymbol m_t}$ addresses the target instance mask, $\mathcal{L}_{\boldsymbol m_o}$ is for the main occluder mask, and $\mathcal{L}_{\boldsymbol m_c}$ is for the main container mask. The ground truth masks for the latter two ($\boldsymbol m_o$ and $\boldsymbol m_c$) are defined to be all-zero whenever there exists no occluder or container respectively, although for class balancing purposes, the loss is also weighted with a factor $\alpha < 1$ for those frames.

Augmentations during training consist of random color jittering (hue, saturation, brightness), random grayscale, random video reversal, random palindromes (\ie playing clips forward and then backward, or vice versa), random horizontal flipping, and random cropping. We do not apply any augmentations at test time.

In Kubric Random, there are many possible objects with available annotations to track. At training time, we assign a difficulty score to every instance (that is visible in the first frame) based on its average occlusion fraction and how much motion it experiences over time. The query is then sampled randomly but non-uniformly, with preference given to the harder to track target objects. At test time, we measure and average metrics over the top four instances with the highest difficulty score per video.
Other datasets (\ie Kubric Containers plus all of Rubric) only have one designated target object per video clip.

In our experiments, we set $(T,H,W)=(30,240,320)$, $\beta=5$, $(\lambda_1, \lambda_2, \lambda_3)=(0.2,0.4,0.4)$, $(\lambda_t,\lambda_o,\lambda_c)=(1.0,0.5,0.5)$, and $\alpha=0.02$. The bootstrap fraction $k$ is a function of time, and decreases linearly from $1$ to $0.15$ during the first 10\% of training. We use the AdamW optimizer and train for 70 epochs, which takes 3 days on 2 NVIDIA RTX A6000 GPUs. Inference (without gradients) happens in 0.27 seconds for a single clip, which corresponds to roughly 110 FPS.

\section{More Qualitative Results}

Please see Figures~\ref{fig:more_succ}, \ref{fig:more_base}, and \ref{fig:more_fail}, as well as \href{https://tcow.cs.columbia.edu/}{tcow.cs.columbia.edu} for videos along with explanations. We recommend viewing the project webpage in a modern browser.

\subsection{Differentiating containers from occluders}

Distinguishing occlusion from containment can be challenging, especially if a potential container is itself responsible for merely occluding but not containing a target object.
One aspect of our TCOW Rubric Office benchmark therefore analyses the interesting scenario where we attempt to trick the model into confusing containment with occlusion. We evaluate this in Figure \ref{fig:confuse}, which illustrates that the TCOW network capitalizes on motion cues, and not (only) object category information.

\end{document}